\documentclass{article}

\usepackage{PRIMEarxiv}

\usepackage[utf8]{inputenc} 
\usepackage[T1]{fontenc}    
\usepackage{hyperref}       
\usepackage{url}            
\usepackage{booktabs}       
\usepackage{amsfonts}       
\usepackage{nicefrac}       
\usepackage{microtype}      
\usepackage{fancyhdr}       
\usepackage{graphicx}       
\graphicspath{{media/}}     
\usepackage{amssymb}
\usepackage{amsmath}
\DeclareMathOperator*{\argmax}{arg\,max}
\DeclareMathOperator*{\argmin}{arg\,min}
\usepackage{multirow}
\usepackage{caption, subcaption}
\usepackage{tikz}
\usepackage{tikz-3dplot}
\usetikzlibrary{bayesnet}
\usepackage[ruled, linesnumbered]{algorithm2e}
\usepackage{mathtools}

\title{Simulation-based Bayesian inference \linebreak for multi-fingered robotic grasping
}

\author{
  Norman Marlier \\
  University of Liège \\
  \texttt{norman.marlier@uliege.be} \\
   \And
  Olivier Brüls \\
  University of Liège \\
  \texttt{o.bruls@uliege.be} \\
  \And
  Gilles Louppe \\
  University of Liège \\
  \texttt{g.louppe@uliege.be} \\
}

\begin{document}
\maketitle

\begin{abstract}
Multi-fingered robotic grasping is an undeniable stepping stone to universal picking and dexterous manipulation. 
Yet, multi-fingered grippers remain challenging to control because of their rich nonsmooth contact dynamics or because of sensor noise.
In this work, we aim to plan hand configurations by performing Bayesian posterior inference through the full stochastic forward simulation of the robot in its environment, hence robustly accounting for many of the uncertainties in the system.
While previous methods either relied on simplified surrogates of the likelihood function or attempted to learn to directly predict maximum likelihood estimates,  we bring a novel simulation-based approach for full Bayesian inference based on a deep neural network surrogate of the likelihood-to-evidence ratio.
Hand configurations are found by directly optimizing through the resulting amortized and differentiable expression for the posterior.
The geometry of the configuration space is accounted for by proposing a Riemannian manifold optimization procedure through the neural posterior.
Simulation and physical benchmarks demonstrate the high success rate of the procedure.
\end{abstract}


\label{sec:introduction}
Two of the grand challenges for the deployment of robots in warehouses, assembly lines or homes are universal picking, \textit{i.e} the ability to robustly grasp a large variety of objects in diverse environments, and dexterous manipulation, \textit{i.e} the ability to manipulate objects to perform useful actions. 
Multi-fingered robotic grasping represents a promising avenue towards these objectives and has the potential to greatly improve and facilitate human-machine interactions.
However, due to the wide variety of object shapes, sensor noise and nonsmooth contact dynamics, multi-fingered grippers remain challenging to control. 
In addition, multi-fingered grasps entail high dimensional configuration spaces compared to two-fingered grasps, making them difficult to optimize and plan.

Early analytical approaches~\cite{Prattichizzo2008,ferrari1992planning} for planning multi-fingered grasps rely on force analysis where a metric is optimized based on the laws of mechanics and 3D models.
These analytical methods require the accurate knowledge of many model and world parameters, which is practically difficult to achieve in real settings.
By contrast, learning-based methods are getting more and more established because they can extract useful information from raw sensor data~\cite{kleeberger2020survey}.
These methods typically make use of machine learning or deep learning models for learning a mapping between grasp configurations and their success. 
This strategy has been applied on top-down grasping with a parallel jaw gripper and has demonstrated excellent results~\cite{Mahler-RSS-17, kalashnikov2018scalable, viereck2017learning, pinto2016supersizing, morrison2018closing}, including its extension to 6 DOF~\cite{ten2017grasp, mousavian20196}.

In this work, we consider the more challenging setting of robust multi-fingered grasping plans, including the 6 DOF grasp poses and the finger configuration.
By framing the problem as an inference task, we demonstrate the generic usefulness and applicability of likelihood-free Bayesian inference algorithms to difficult robotic tasks such as multi-fingered grasping considered here.
We summarize our contributions as follow:
\begin{itemize}
    \item We bring simulation-based Bayesian inference methods~\cite{cranmer2020frontier} to multi-fingered grasping. By learning a model for the likelihood-to-evidence ratio and using an analytical prior, we derive an amortized and differentiable posterior for the hand configurations.
    \item We make use of Riemannian manifold optimization to deal with the nonlinearity of the configuration space, in particular of the 3D rotation group tied to the hand pose.
    \item We validate our approach on a realistic and challenging 3-finger gripper setup, through both simulation and real-world physical benchmarks. Results demonstrate its high success rate.
\end{itemize}


\section{Problem statement}

\label{sec:statement}

\begin{figure}
\centering
    \resizebox{\linewidth}{!}{
    \begin{subfigure}{.5\textwidth}
    \centering
    \resizebox{\linewidth}{!}{\input{frame_diagram}}
    \caption{}
    \label{fig:sim_scene}
    \end{subfigure}
    \begin{subfigure}{.5\textwidth}
    \centering
    \resizebox{\linewidth}{!}{\includegraphics{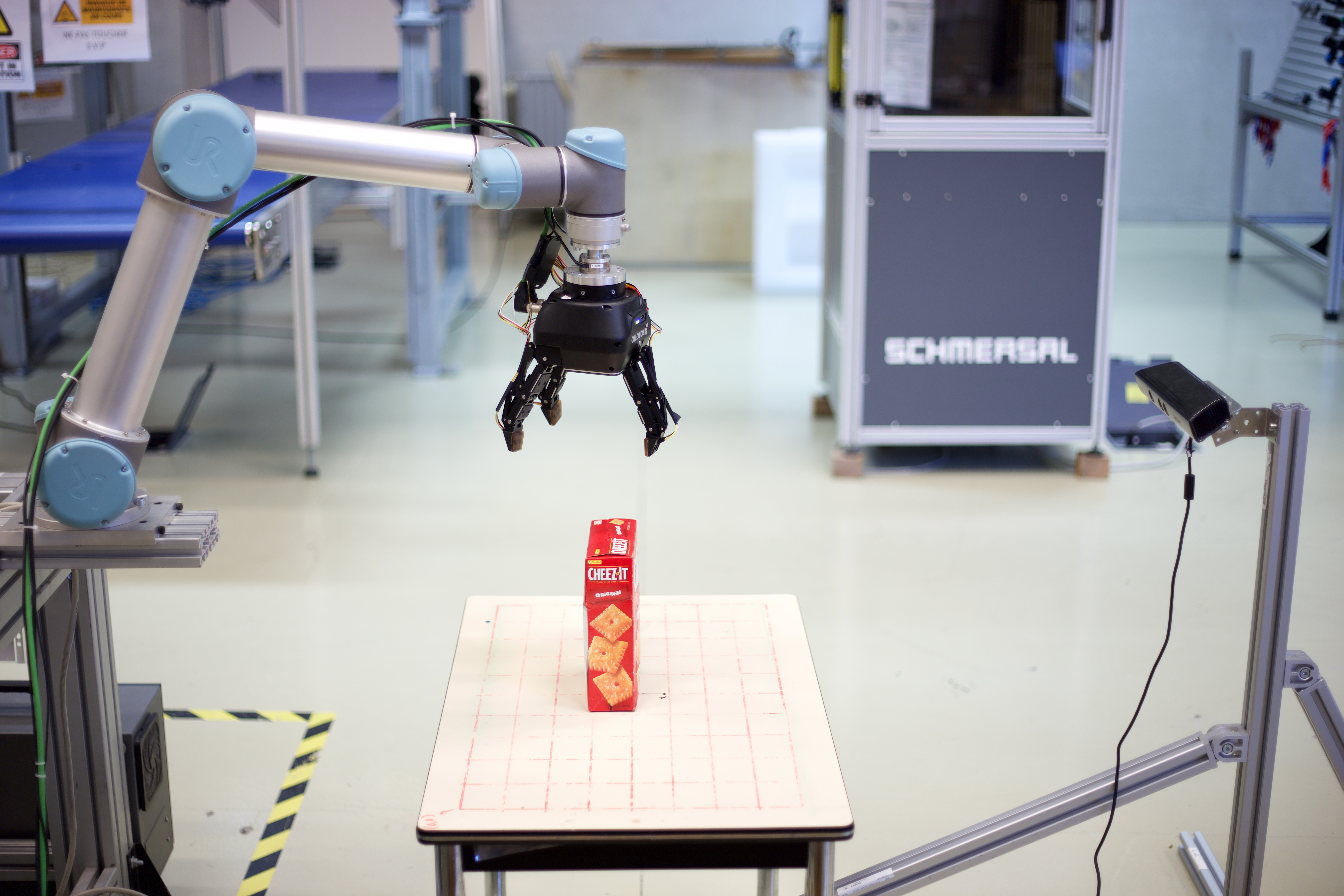}}
    \caption{}
    \label{fig:real_scene}
    \end{subfigure}}
    \caption{Our benchmark scene for multi-fingered grasping. 
The pose of the hand $(\mathbf{x}, \mathbf{R})$ is defined in the local object frame. The depth camera produces an image $i$ of the scene. }
    \label{fig:frame_diagram}
\end{figure}

\paragraph{Description}
We consider the problem of grasping a rigid and stable body with a multi-fingered gripper, as illustrated in Fig.~\ref{fig:frame_diagram}.
The object $\mathcal{O}$ is modelled as a 3D surface mesh and its centroid stands on a table at a location $(x_{\mathcal{O}}, y_{\mathcal{O}}, z_{\mathcal{O}})$ with a rotation $\varphi_{z,\mathcal{O}}$ around the $z$-axis in the world reference frame $\mathcal{F}_{W}$.
We refer to its 2D pose $(x_\mathcal{O}, y_\mathcal{O}, \varphi_{z,\mathcal{O}})$ as $\mathbf{p}_{\mathcal{O}} \in \mathbb{R}^{2}\times \text{SO(2)}$. 
The hand configuration $\mathbf{h} \in \mathcal{H} = \mathbb{R}^{3}\times \text{SO(3)} \times \mathcal{G}$ is defined as the combination of the pose $(\mathbf{x},\mathbf{R}) \in \mathbb{R}^{3}\times \text{SO(3)}$ of the hand and the type $g \in \mathcal{G} = \{ \text{basic}, \text{wide}, \text{pinch} \}$ of the grasp.
The hand pose $(\mathbf{x}, \mathbf{R})$ is defined with respect to the world frame coordinate.
The robot evolves in a 3D workspace  observed with a fixed depth camera producing images $i \in \mathcal{I}$.
The goal is to find a robust hand configuration $\mathbf{h^{*}}$  with respect to a given binary metric $S = \{0, 1\}$, where $S=1$ indicates a successful grasp.

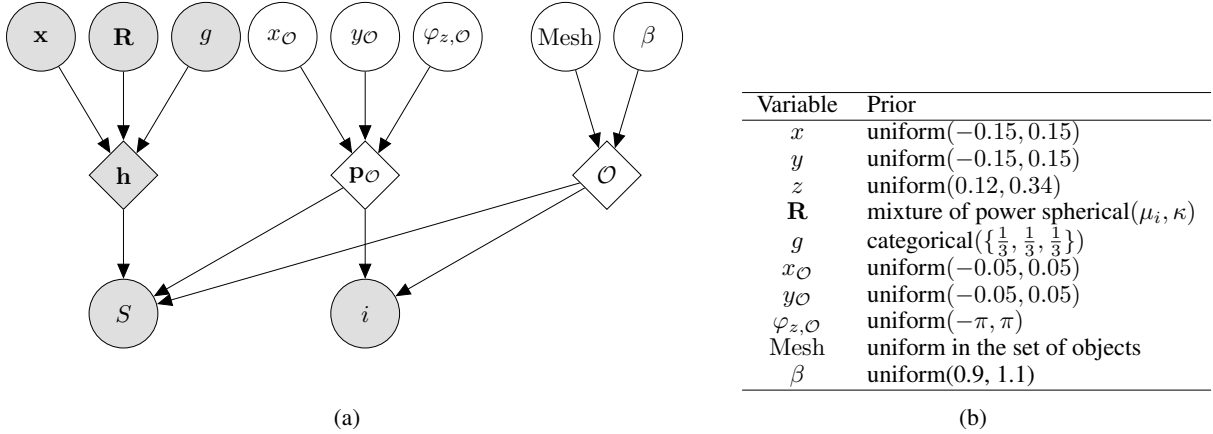
\begin{figure}
    \begin{subfigure}[b]{.55\linewidth}
    \centering
    \resizebox{\linewidth}{!}{\begin{tikzpicture}[scale=1]
  
  
  \node[det, fill=gray!25, minimum size=1cm] (hv) {$\mathbf{h}$};
  \node[obs, above=1cm of hv, minimum size=1cm] (R) {$\mathbf{R}$};
  \node[obs, above=1cm of hv, xshift=-1.2cm, minimum size=1cm] (x) {$\mathbf{x}$};
  \node[obs, above=1cm of hv, xshift=1.2cm, minimum size=1cm] (g) {$g$};
  
  \edge{x, R, g}{hv};

  \node[det, right=2.5cm of hv, minimum size=1cm] (pO) {$\mathbf{p_{\mathcal{O}}}$};
  \node[latent, above=1cm of pO, xshift=-1.2cm, minimum size=1cm] (xO) {$x_{\mathcal{O}}$};
  \node[latent, above=1cm of pO, minimum size=1cm] (yO) {$y_{\mathcal{O}}$};
  \node[latent, above=1cm of pO, xshift=1.2cm, minimum size=1cm] (rO) {$\varphi_{z, \mathcal{O}}$};
  
  \edge{xO, yO, rO}{pO};
  
  \node[det, right=2.5cm of pO, minimum size=1cm] (Ov) {$\mathcal{O}$};
  \node[latent, above=1cm of Ov, xshift=-0.6cm, minimum size=1cm] (M) {$\mathrm{Mesh}$};
  \node[latent, above=1cm of Ov, xshift=0.6cm, minimum size=1cm] (scale) {$\beta$};
  \edge{M, scale}{Ov};

  \node[obs, below=1cm of hv, minimum size=1cm] (S) {$S$};
  \node[obs, below=1cm of pO, minimum size=1cm]  (i) {$i$};

  \edge{hv, pO, Ov}{S};
  \edge{pO, Ov}{i};
  
\end{tikzpicture}}
    \vspace{0.5em}
    \caption{}
    \label{fig:graphical_model}
\end{subfigure}
\begin{subfigure}[b]{.45\linewidth}
\centering
\resizebox{0.85\linewidth}{!}{
\begin{tabular}{cl}
    \hline
    Variable    &  Prior   \\
    \hline
    $x$ &  $\text{uniform}(-0.15,0.15)$ \\
    $y$ &  $\text{uniform}(-0.15,0.15)$ \\
    $z$ &  $\text{uniform}(0.12,0.34)$ \\
    $\mathbf{R}$ & $\text{mixture of power spherical}(\mathbf{\mu}_{i}, \kappa)$ \\
    $g$ & $\text{categorical}(\{\frac{1}{3},\frac{1}{3},\frac{1}{3}\})$ \\
    $x_{\mathcal{O}}$ & $\text{uniform}(-0.05,0.05)$ \\
    $y_{\mathcal{O}}$ & $\text{uniform}(-0.05,0.05)$ \\
    $\varphi_{z, \mathcal{O}}$ & $\text{uniform}(-\pi,\pi)$ \\
    $\mathrm{Mesh}$ & $\text{uniform in the set of objects}$ \\
    $\beta$ & uniform(0.9, 1.1)\\
    \hline
    \end{tabular}}
    \caption{}
    \label{tab:prior_distribution}
\end{subfigure}
\caption{(a) Probabilistic graphical model of the environment. Gray nodes correspond to observed variables and white nodes to unobserved variables. (b) Prior distributions.}
\end{figure}

\paragraph{Probabilistic modeling}
We model the scene and the grasping task according to the Bayesian network shown in Fig.~\ref{fig:graphical_model}.
The variables $S, i, \mathbf{h}, \mathcal{O}$ and  $\mathbf{p}_{\mathcal{O}}$ are modelled as random variables in order to capture the noise in the robot or in the depth camera, as well as our prior beliefs about the hand configuration, the object or its pose. 
The structure of the Bayesian network is motivated by the fact that $\mathbf{h}$, $\mathcal{O}$ and $\mathbf{p}_\mathcal{O}$ are independent, while $S$ is dependent on $\mathbf{h}$, $\mathcal{O}$ and $\mathbf{p}_\mathcal{O}$ and $i$ is dependent on $\mathcal{O}$ and $\mathbf{p}_\mathcal{O}$. This structure also enables a direct and straightforward way to generate data: $\mathbf{h}$, $\mathcal{O}$ and $\mathbf{p}_\mathcal{O}$ are sampled from their respective prior distributions while $S$ and $i$ can be generated using forward physical simulators for the grasping and the camera.

The prior distribution $p(\mathbf{x})$ of the spatial position is uniformly distributed between the extreme values $\mathbf{x}_{\text{lim}}=(x_{\text{low}},y_{\text{low}}, z_{\text{low}}, x_{\text{high}},y_{\text{high}},z_{\text{high}})$, chosen to be within the range of physical dimensions of the gripper and the biggest object. It emphasizes our ignorance about interesting regions of space for grasping. The rotation $\mathbf{R}$ is parameterized with a quaternion. A quaternion $\mathbf{q}$ is an element of the quaternion group $\mathbb{H}$, in the form $\mathbf{q}= q_{0}\mathbf{1} + q_{1}\mathbf{i} + q_{2}\mathbf{j} + q_{3}\mathbf{k}  = (q_{0}, q_{1}, q_{2}, q_{3})^{T}$ with $(q_{0}, q_{1}, q_{2}, q_{3})^{T} \in \mathbb{R}^{4}$ and $\mathbf{i}^{2}=\mathbf{j}^{2}=\mathbf{k}^{2}=\mathbf{ijk}=-1$. The conjugate $\mathbf{\bar{q}}$ of quaterion $\mathbf{q}$ is given by $\mathbf{\bar{q}}:= q_{0}\mathbf{1} - q_{1}\mathbf{i} - q_{2}\mathbf{j} - q_{3}\mathbf{k}$. A unit quaternion, called \textit{versor}, $\mathbf{q}_{1} \in \mathbb{H}_{1}$  has a unit norm defined as $\|\mathbf{q}\| = \sqrt{ \mathbf{q}\mathbf{\bar{q}}}=1$. They give a more compact representation than rotation matrices and avoid gimbal lock and singularities. Unit quaternions can be identified with the elements of a hyperspherical manifold $\mathbb{S}^{3}$ embedded into $\mathbb{R}^{4}$. Moreover, $\mathbb{S}^{3}$ is a double covering of $\text{SO(3)}$, meaning that antipodal points $\pm\mathbf{q}$ represent the same rotation, which implies that $p(\mathbf{q};\cdot)=p(-\mathbf{q};\cdot)$. We define the prior $p(\mathbf{q})$ as a mixture of \textit{power-spherical} distributions \cite{de2020power} with 4 modes $\mathbf{\mu}_{i}$. Each mode is a mixture that satisfies $p(\mathbf{q};\cdot)=p(-\mathbf{q};\cdot)$. In total, we have
\begin{equation}p(\mathbf{q}) = \frac{1}{N} \sum_{i=1}^{N=4}\frac{\text{PowerSpherical}(\mathbf{q}; \mathbf{\mu}_{i}, \kappa)}{2} + \frac{\text{PowerSpherical}(\mathbf{q}; -\mathbf{\mu}_{i}, \kappa)}{2}.
\end{equation}
These modes $\mathbf{\mu}_{i}$ encode information about the orientation of the gripper and share the same concentration factor $\kappa=30$.
To grasp an object, the gripper point toward the table and thus toward the object -- an informed prior which indeed results in sufficiently many successful grasps. 
We then define four rotations, separated by a rotation of $\frac{\pi}{2}$ around the $z$-axis (see Fig.~\ref{fig:prior_quat_mode} in Appendix~\ref{appendix:distribution}).
In this way, our prior covers a large part of the rotation space and is sufficiently informative by contrast to a uniform prior over the unit sphere $\mathbb{S}^{3}$.
The grasp type $g$ is uniformly distributed between the three types basic, wide and pinch. These three modes modulates the spacing between the fingers in the opposite side of the thumb. Finally, $p(\mathbf{h}) = p(\mathbf{x})p(\mathbf{R})p(g)$.

The prior $p(\mathcal{O})=p(\mathrm{Mesh})p(\beta)$ is a discrete uniform distribution over a fixed set of object meshes and a uniform distribution for the scaling factor $\beta$. 
Finally, the prior $p(\mathbf{p}_{\mathcal{O}})$ captures our belief that the object can be everywhere on the table with any rotation around the vertical axis. For this reason, uniform distributions are used for all three parameters $x_{\mathcal{O}},y_{\mathcal{O}}, \varphi_{z, \mathcal{O}}$.
Table~\ref{tab:prior_distribution} summarizes the prior distributions.

Given our probabilistic graphical model, we finally formulate the problem of grasping as the Bayesian inference of the hand configuration $\mathbf{h}^{*}$ that is a posteriori the most likely given a successful grasp and an observation $i$. That is, we are seeking for the maximum a posteriori (MAP) estimate
\begin{equation}
\label{eq:map}
\mathbf{h}^{*} = \argmax_{\mathbf{h}}~p(\mathbf{h}|S=1, i).
\end{equation}

\section{Likelihood-free Bayesian inference for multi-fingered grasping}
\label{sec:method}

\subsection{Density ratio estimation}
From the Bayes's rule, the posterior of the hand configuration is
\begin{equation}
\label{eq:proba_cond}
\begin{split}
p(\mathbf{h}|S, i)  = \frac{p(S,i| \mathbf{h})}{p(S,i)}p(\mathbf{h}). \\
\end{split}
\end{equation}
The likelihood function $p(S, i|\mathbf{h})$ and the evidence $p(S, i)$ are both intractable, which makes standard Bayesian inference procedures such as Markov chain Monte Carlo unusable. 
However, drawing samples from forward models remains feasible with physical simulators, hence enabling likelihood-free Bayesian inference algorithms. 

First, we express the likelihood-to-evidence ratio as a product of two individual ratios,
\begin{align}
    r(S, i|\mathbf{h}) &= \frac{p(S, i|\mathbf{h})}{p(S, i)}= \frac{p(S|\mathbf{h})}{p(S)}  \frac{p(i|S, \mathbf{h})}{p(i|S)}= r(S|\mathbf{h}) r(i|S, \mathbf{h}).
    \label{eq:ratio_decomposition}
\end{align}
By adapting the approach described in \cite{pmlr-v119-hermans20a, Brehmer:2019jyt} for likelihood ratio estimation, we train two neural network classifiers $d_\phi$ and $d_\theta$ that we will use to approximate $r(S|\mathbf{h})$ and $r(i|S, \mathbf{h})$.
The first network $d_\phi$ is trained to distinguish positive tuples $(S, \mathbf{h})$ (labeled $y=1$) sampled from the joint distribution $p(S, \mathbf{h})$ against negative tuples (labeled $y=0$) sampled from the product of marginals $p(S)p(\mathbf{h})$. The Bayes optimal classifier $d^{*}(S,\mathbf{h})$ that minimizes the cross-entropy loss is given by
\begin{equation}
\label{eq:discriminator}
d^{*}(S, \mathbf{h}) = \frac{p(S, \mathbf{h} )}{p(S)p(\mathbf{h})+ p(S, \mathbf{h})},
\end{equation}
which recovers the likelihood ratio $r(S|\mathbf{h})$ as
\begin{equation}
    \label{eq:d_to_r}
    \begin{split}
    \frac{d^{*}(S,\mathbf{h})}{1-d^{*}(S,\mathbf{h})} & = \frac{p(S,\mathbf{h})}{p(S)p(\mathbf{h})} = \frac{p(S|\mathbf{h})}{p(S)}.
    \end{split}
\end{equation}
Therefore, by modelling the classifier with a neural network $d_\phi$ trained on the binary classification problem, we obtain an approximate but amortized and differentiable likelihood ratio  
\begin{equation}
    \hat{r}(S|\mathbf{h}) = \frac{d_\phi(S,\mathbf{h})}{1-d_\phi(S,\mathbf{h})}.
\end{equation}
The second network $d_\theta$ is trained similarly, over positive tuples $(i, \mathbf{h})$ (labeled ($y=1$) sampled from the conditional joint distribution $p(i, \mathbf{h}|S=1)$ against negative tuples $(i, \mathbf{h})$ (labeled $y=0$) sampled from the product of marginals $p(i|S=1)p(\mathbf{h}|S=1)$. Using the same likelihood ratio trick, we obtain 
\begin{equation}
    \hat{r}(i|S=1, \mathbf{h}) = \frac{d_\theta(i, \mathbf{h})}{1-d_\theta(i, \mathbf{h})}.
\end{equation}
Finally,  the likelihood ratios are combined with the prior to approximate the posterior as
\begin{equation}
\hat{p}(\mathbf{h}|S=1, i) =  \hat{r}(i|S=1, \mathbf{h})\hat{r}(S=1|\mathbf{h}) p(\mathbf{h}),
\end{equation}
which enables immediate posterior inference despite the initial intractability of the likelihood function $p(S, i|\mathbf{h})$ and of the evidence $p(S, i)$.

\begin{figure}
        \centering
        \input{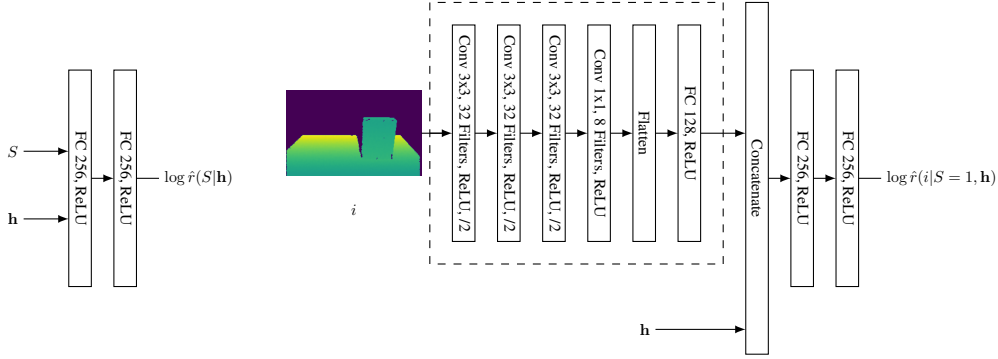}
        \label{fig:image_estimator}
    \caption{Neural network architectures of the classifiers $d_\phi$ and $d_\theta$ used to respectively approximate the likelihood ratios $r(S|\mathbf{h})$ and $r(i|S=1,\mathbf{h})$. }
    \label{fig:pipeline_architecture}
\end{figure}

The neural network classifiers $d_\phi$ and $d_\theta$ are architectured as in Fig.~\ref{fig:pipeline_architecture}. 
In $d_\theta$, the camera image $i$ of size $640\times 480\times 1$ is pre-processed  by scaling the depths in the interval $\{0\} \cup [0.45, 1]$ and resized to $256\times 160\times 1$. Then, $i$ is fed to a convolutional network made of four convolutional layers followed by a fully connected layer and which goal is to produce a vector embedding of the image.
The image embedding and $\mathbf{h}$ are then both fed to a subsequent network made of 2 fully connected layers. 
The hand configuration $\mathbf{h}$ enters the neural network as a $1\times13$ vector where the rotation matrix $\mathbf{R}$ is flattened \cite{murphy2021implicit, zhou2019continuity} and the grasp type $g$ is passed through an embedding.
In $d_\phi$, both $S$ and $\mathbf{h}$ are directly fed to a network made of 2 fully connected layers. 
The parameters $\phi$ and $\theta$ are optimized by following Algorithm~\ref{alg:learning_procedure} (Appendix~\ref{appendix:algorithms}), using Adam as optimizer.

Finally, we note that the factorization of the likelihood-to-evidence ratio forces the two ratio estimators to focus their respective capacity on the information brought by $S$ and $i$. Because of the high discriminative power of $S$, training instead a single ratio taking both $S$ and $i$ as inputs would indeed lead to an estimator that usually discards the smaller information brought in $i$.

\subsection{Maximum a posteriori estimation}
\label{subsec:optimization}

Due to the intractability of the likelihood function and of the evidence,
Eq.~(\ref{eq:map}) cannot be solved analytically nor numerically. 
We  rely instead on the  approximation given by the likelihood-to-evidence ratio $\hat{r}$ to find an approximation of the maximum a posteriori (MAP) estimate as
\begin{align}
    \hat{\mathbf{h}}^{*} &= \argmax_{\mathbf{h}}\hat{r}(S=1,i|\mathbf{h})p(\mathbf{h}) \\
    &= \argmin_{\mathbf{h}} -\log \hat{r}(S=1,i|\mathbf{h})p(\mathbf{h})
    \label{eq:approximate_map},
\end{align}
which we solve using gradient descent. 
For a given $g$, the gradient of Eq.~(\ref{eq:approximate_map}) decomposes as
\begin{equation}
\begin{aligned}
\label{eq:euclidean_grad}
    -\nabla_{(\mathbf{x},\mathbf{R})}\log \hat{r}(S,i|\mathbf{h})p(\mathbf{h}) =  -\nabla_{(\mathbf{x},\mathbf{R})}\log &\,\hat{r}(S,i|\mathbf{h}) - \nabla_{(\mathbf{x},\mathbf{R})}\log p(\mathbf{h}).
\end{aligned}
\end{equation}
Our closed-form prior $p(\mathbf{h})$ has analytical gradients. In fact, uniform distributions are set to have null gradient everywhere in the domain. Therefore, $\nabla_{\mathbf{x}}p(\mathbf{h}) = \mathbf{0}$. By contrast, $p(\mathbf{R})$ is a weakly informative prior and has a non null gradient from the power spherical distribution. Its derivative with respect to $\mathbf{q}$ is
\begin{equation}
\label{eq:grad_power_spherical}
\begin{split}
\nabla_{\mathbf{q}}p(\mathbf{q};\mu, \kappa) &=  C(\kappa)\kappa(1+\mu^{T}\mathbf{q})^{\kappa-1}\nabla_{\mathbf{q}}(1+\mu^{T}\mathbf{q})\\
 &= C(\kappa)\kappa\mathbf{\mu}(1+\mu^{T}\mathbf{q})^{\kappa-1},
\end{split}
\end{equation}
where $C(\kappa)$ is the normalization term.
Since the likelihood-to-evidence ratio estimator $\hat{r}$ is modelled by a neural network, it is fully differentiable with respect to its inputs and its gradients can be computed by automatic differentiation. 
However, not all variables of the problem are Euclidean variables and naively performing gradient descent would violate our geometric assumptions. 
Let us consider a variable $\mathcal{Z}$ on the smooth Riemannian manifold $\mathcal{M}=\mathbb{R}^{3} \times \text{SO(3)}$  with tangent space $\mathcal{T}_{\mathcal{Z}}\mathcal{M}$ and a function $f : \mathcal{M} \rightarrow \mathbb{R}$. Since SO(3) is embedded in the set of $3\times 3$ matrices $\mathbb{R}^{3\times 3}$, $f$ can be evaluated on $\mathbb{R}^{3} \times \mathbb{R}^{3\times3} $, leading to the definition of the Euclidean gradients $\nabla f(\mathcal{Z}) \in \mathbb{R}^{3} \times \mathbb{R}^{3\times3}$. In turn, these Euclidean gradients can be transformed into their Riemannian counterparts $\text{grad}f(\mathcal{Z})$ via orthogonal projection $\mathbf{P}_{\mathcal{Z}}$ into the tangent space $\mathcal{T}_{\mathcal{Z}}\mathcal{M}$ \cite{absil2009optimization, hu2019brief}. Therefore,
\begin{equation}
        \text{grad} f(\mathcal{Z}) =  \mathbf{P}_{\mathcal{Z}}(\nabla f(\mathcal{Z}))
\end{equation}
where the orthogonal projection onto $\mathbb{R}^{3}$ is the identity $\mathbb{I}_{3}$ and the orthogonal projection onto SO(3) at $\xi \in \text{SO(3)}$ is $\xi\text{skew}(\xi^{T}\nabla f(\xi))$ where $\text{skew}(A) \coloneqq \frac{1}{2}(A-A^{T})$. Thus, we can solve Eq.~(\ref{eq:approximate_map}) by projecting Euclidean gradients of Eq.~(\ref{eq:euclidean_grad}) to the tangent space $\mathcal{T}_{\mathcal{Z}}\mathcal{M}$ and plug them to a manifold optimization procedure. In our experiments, we use the geometrical conjugate gradient method \cite{absil2009optimization} implemented in Pymanopt~\cite{townsend2016pymanopt} to perform 20 optimization steps and we scan for the best value of $g$. 
For completeness, the full optimization algorithm is provided in Algorithm~\ref{alg:optimization_procedure} (Appendix~\ref{appendix:algorithms}).

\section{Experiments}
\label{sec:experiment}

For training, we use 19 objects from the YCB data \cite{7251504} (see Fig.~\ref{fig:training_set} in Appendix~\ref{appendix:objects}) together with 5 objects from the ShapeNet dataset \cite{shapenet2015}, for a total of 24 types of objects. We selected a diverse range of objects compatible with the geometry of our gripper.
In simulation, the success rate was evaluated on the 19 objects used for training, as well as on 
5 new unseen objects from the YCB data (see Fig.~\ref{fig:testing_set} in Appendix~\ref{appendix:objects}).
Only the 19 objects from YCB are used in the real setup.

\subsection{Data generation}
\paragraph{Grasp generative model}
A physical simulator is used to sample from $p(S|\mathbf{h}, \mathcal{O}, \mathbf{p}_{\mathcal{O}})$. Hand configurations, objects and object poses are sampled from their priors $p(\mathbf{h})$, $p(\mathcal{O})$ and $p(\mathbf{p}_{\mathcal{O}})$, and  are then submitted to a lift test. First, a planner generates a trajectory in the joint space to bring the gripper to the hand configuration $\mathbf{h}$. If the pose is not reachable, the test fails. Otherwise, the gripper closes its fingers until contact with the object or the gripper itself. Then, the robot lifts possibly the object to a given height. If the object is held in the gripper, the grasp is considered as successful. Simulations are performed using Pybullet~\cite{coumans2020}.
We use volumetric hierarchical approximate decomposition to get convex meshes of objects from \textit{obj} files for collision detection \cite{mamou2009simple}.

\paragraph{Sensor generative model}
The sensor generative model $p(i|\mathbf{p}_{\mathcal{O}}, \mathcal{O})$ is implemented in Pybullet, with an approach similar to the Blensor sensor framework \cite{gschwandtner2011blensor} used to render depth images from a Kinect model sensor. Simulating a structured-light sensor allows a better transfer to the real setup. 
Objects and poses are sampled from their priors, $\mathcal{O} \sim p(\mathcal{O}), \mathbf{p}_{\mathcal{O}}\sim p(\mathbf{p}_{\mathcal{O}})$. Then, the object is placed and an image $i \sim p(i|\mathbf{p}_{\mathcal{O}}, \mathcal{O})$ is generated.

\paragraph{Domain randomization for sim-to-real transfer}
Generative models in simulation differ from their real world counterparts due to incorrect physical modelling and inaccurate physical parameters. This \textit{reality gap} may lead to failures from our model because the synthetic data and the real data distributions are different. To overcome this issue, we use \textit{domain randomization} \cite{tobin2017domain} with nuisance parameters on the position and the orientation of the camera, the minimum and maximum distance of the depth, the field of view, and the coefficient of lateral and spinning frictions $\mu$ and $\gamma$.
Domain randomization avoids the precise calibration of both the grasp and the image simulators, which can be very difficult and costly.  
We use uniform distributions for the nuisance parameters which are difficult to measure with $\pm 2\%$ error and Gaussian distributions for easily measurable parameters. For the orientation of the camera, a multivariate normal variable $\eta \sim \mathcal{N}(\mathbf{0}, \Sigma), \Sigma=\text{diag}(\sigma_{\alpha}=0.002, \sigma_{\beta}=0.01, \sigma_{\gamma}=0.002)$ is drawn and then mapped to $\text{SO}(3)$ using the exponential map.

\subsection{Simulation benchmarks}
We evaluate the performance of our approach incrementally, adding algorithmic components one by one to assess their respective marginal benefits.
For each inference strategy,  we estimate the success rate over $1000$ grasping attempts for randomly sampled objects and camera images. Nuisances parameters are resampled in simulation when evaluating the success.
Our general results are summarized in Table~\ref{tab:sim_metric}, while supplementary results for each individual category of objects can be found in Appendix~\ref{appendix:supplementary-results}.
We first report results for strategies maximizing the (conditional) densities $p(\mathbf{h}|\cdot)$ of hand configurations.
Optimizing for the maximum a priori estimate $\mathbf{h} = \argmax p(\mathbf{h})$, without conditioning on success or an observation of the scene, leads to a very low success rate of $0.6\%$. 
As expected, these results are too poor to be usable but they should underline the informativeness of the prior. Sampling hand configurations from a uniform prior would instead result in a much smaller success rate, by about one order of magnitude (less than $0.1\%$).
When conditioning on the expected success $S=1$, performance increases to $44\%$.
Taking both the expected success $S=1$ and the image $i$ into account and following the inference methodology proposed in Section~\ref{sec:method}, leads to an even larger success rate of $71\%$ for the maximum a posteriori estimates. For the 5 new objects, we reach a comparable success rate of $75\%$, which demonstrates the good generalization of the approach.
In comparison, had the properties $\mathcal{O}$, $\mu$, and $\beta$ of the object been perfectly known, the success rate would reach $85\%$, which shows that the convolutional network manages to extract most of the relevant information from the observation $i$.
Table~\ref{tab:sim_metric} also reports results for maximum likelihood estimates, achieving success rates of $43\%$, $64\%$ and $80\%$ when maximizing the likelihoods $p(S=1|\mathbf{h})$, $p(S=1,i|\mathbf{h})$, and $p(S=1, \mathcal{O}, \mu, \beta|\mathbf{h})$ respectively. Note that maximizing $p(S=1|i, \mathbf{h})$ and $p(S=1| \mathcal{O}, \mu, \beta,\mathbf{h})$ would give the same result since $i$ and $\mathcal{O}, \mu, \beta$ are independent from $\mathbf{h}$.
Our informative prior can explain the difference in success rates between the MAP and the MLE estimates and motivates the use of a Bayesian approach.

\begin{figure}
    \centering
    \resizebox{0.70\linewidth}{!}{\includegraphics{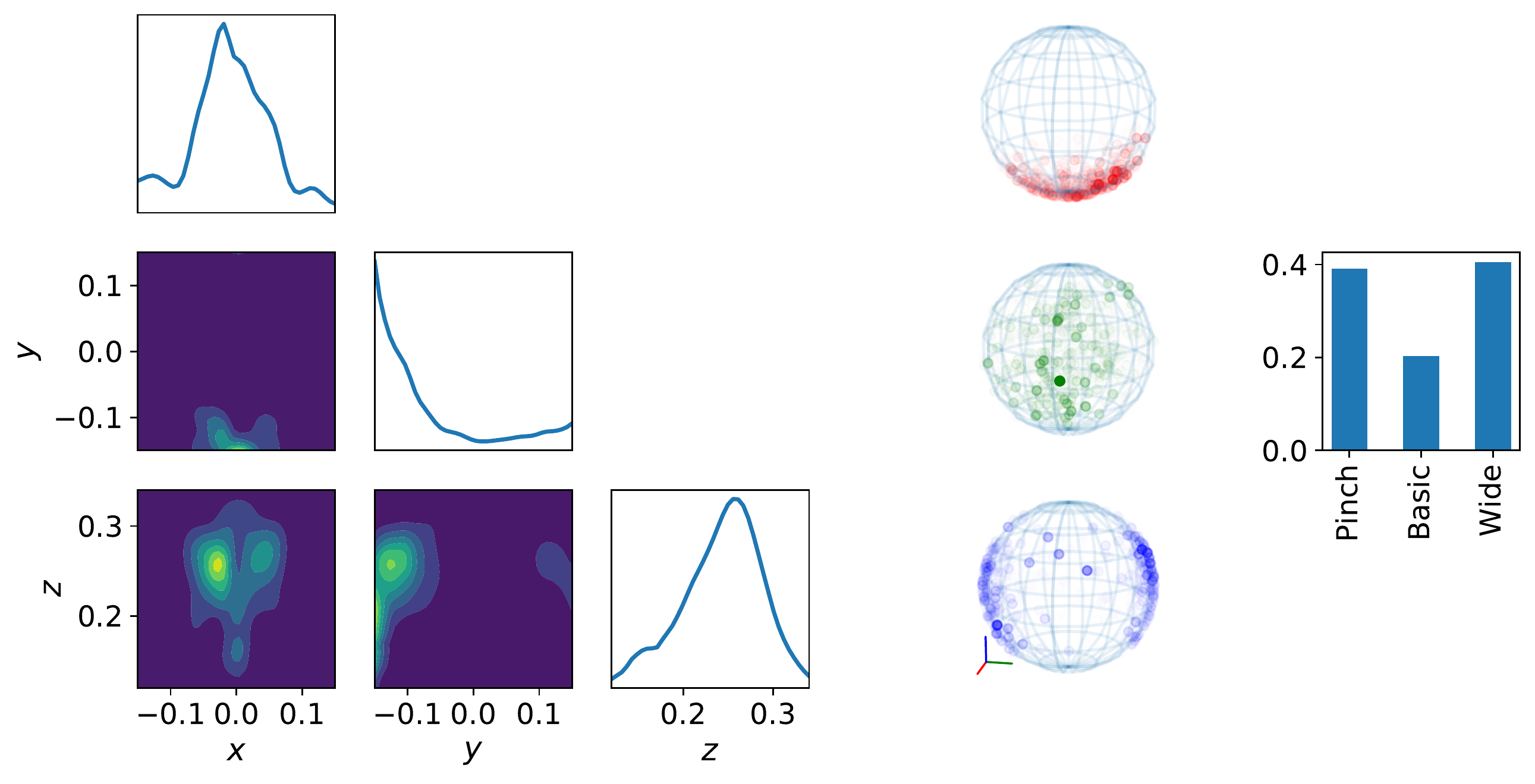}}
    \caption{Posterior $p(\mathbf{h}|S=1, i)$ for the setup in Fig.~\ref{fig:sim_scene} with the box centered at $(0, 0)$. (Left) Posterior $p(\mathbf{x}|S=1, i)$, (Middle) Posterior $p(\mathbf{R}|S=1, i)$, (Right) Posterior $p(g|S=1, i)$.}
    \label{fig:posterior}
\end{figure}

Not only our framework can be used for grasp planning, it also provides immediate access to the full posterior $p(\mathbf{h}|S=1, i)$ of hand configurations.
As an illustrating example, we extract the marginal posterior densities $p(\mathbf{x}|S=1, u), p(\mathbf{R}|S=1, i)$ and $p(g|S=1, i)$ for the simulated scene of Fig.~\ref{fig:sim_scene}, with the box centered at $(0, 0)$ without any rotation. 
The resulting posterior is shown in Fig.~\ref{fig:posterior}.
First, $p(\mathbf{x}|S=1, i)$ shows the distribution in space of the hand configuration $\mathbf{h}$. 
The concentration along the $x$-axis and $z$-axis are high, meaning that high density regions are located slightly behind and in front of the box, at a given height related to the geometrical dimensions of the box. 
Concerning the $y$-axis, the posterior fails to capture the symmetry and places all the density at the right of the box. 
Overall, the positions $\mathbf{x}$ which are the most likely to give a successful grasp are on the right corner of the box. 
This is underlined by the posterior of $p(\mathbf{R}|S=1, i)$. The red dots correspond to the density of the $x$-axis, the green dots to the $y$-axis and the blue dots to the $z$-axis. 
The $x$-axis has one mode, directed toward the table, inherited from the prior and slightly deviates to the right. The $y$-axis, however, has only two antipodal modes by contrast to the prior. These modes correspond to the situation in which the fingers are placed on the front surface or the back surface. The $z$-axis can be constructed by taking the cross product between $x$ and $y$. Uncertainties from $x$ and $y$ are propagated, leading to two antipodal modes with lower concentration than $y$. 
Finally, the posterior $p(g|S=1, i)$ for the grasp type indicates a preference towards pinch and wide modes over the basic mode. Pinch mode is preferred when the position $\mathbf{x}$ is far from the right corner while the wide mode is mainly used when $\mathbf{x}$ is located near the right corner.

\begin{table}
\centering
    \resizebox{0.80\linewidth}{!}{
    \begin{tabular}{llcc}
    \hline
    \hline
    Grasping inference strategy & & \multicolumn{2}{c}{Success rate}\\
    \hline
     & & {\it Sim} & {\it Real} \\
     \hline
    \multirow{2}{*}{Prior based} & $\mathbf{h} \sim p(\mathbf{h})$ & $0.8\%$ & - \\
    & $\mathbf{h}=\argmax p(\mathbf{h})$          & $0.6\%$ & - \\
    \hline
    \multirow{2}{*}{Metric based} & $\mathbf{h}=\argmax_{\mathbf{h}}  \hat{p}(S=1| \mathbf{h})$ & $43\%$ & - \\
     & $\mathbf{h}=\argmax_{\mathbf{h}} \hat{p}(\mathbf{h}| S=1)$ & $44\%$  & $46\%$ \\
    \hline
    \multirow{2}{*}{Partial observation based}
    & $\mathbf{h}=\argmax_{\mathbf{h}}  \hat{p}(S=1, i | \mathbf{h})$ & $64\%$$/71\%$ & - \\
    & $\mathbf{h}=\argmax_{\mathbf{h}} \hat{p}(\mathbf{h}| S=1, i)$ & $71\%$$/\mathbf{75}\%$ & $70\%$ \\
    \hline
    \multirow{2}{*}{Full observation based (ideal)} &
     $\mathbf{h}=\argmax_{\mathbf{h}}  \hat{p}(S=1, \mathcal{O}, \mu, \beta| \mathbf{h})$ & $80\%$  & -\\
     & $\mathbf{h}=\argmax_{\mathbf{h}} \hat{p}(\mathbf{h}|S=1, \mathcal{O}, \mu, \beta)$ & $85\%$ & - \\
    \hline
    \hline
    \end{tabular}}
    \vspace{1em}
    \caption{Grasping success rate for various inference strategies of the hand configuration. 
    The success rate obtained by performing Bayesian posterior inference through the full  forward simulation reaches  $71\%$ for objects seen during training and $75\%$ for 5 new objects. In real experiments, the success rate reaches $70\%$. 
    }
    \label{tab:sim_metric}
\end{table}

\subsection{Physical benchmarks}
We carried out experiments with a Robotiq 3-finger gripper attached to a UR5 robotic arm, as shown in Figure~\ref{fig:real_scene}. A Kinect v1 provides depth images and is rigidly linked to the robot base. The robotic arm is controlled in position in the joint space. 
Communications are performed within the ROS framework. 
We calibrate the centre of the table by computing the trajectory in the simulator and then sending it to the real robot. 
We perform 10 trials per object, for a total of 190 grasps. Objects are placed at best at $(x_{\mathcal{O}}=0, y_{\mathcal{O}}=0, \theta_{\mathcal{O}}=0)$. 
As shown in Table~\ref{tab:sim_metric}, our success rate of 70\% is similar than in simulation, which indicates that the simulation-to-reality transfer works well, at least on average. 
These results also demonstrate competitive performance with respect to related works (see Section~\ref{sec:related_work}), although this remains difficult to establish because of the distinct hardware setups.
We observe that failures are mainly behavioral and geometric.
Behavioral failures arise when the simulation does not model the physics correctly. For example, in the real setup, the bowl slides on the table when the gripper closes its fingers, while in simulation, the bowl is just lifted. We could reduce these errors by using a more accurate simulator. Geometric failures arise when there is a shift in the location or in the orientation of the object. Most of the time, the robot either collides with the object or misses it for smaller objects. These failures could be avoided using a more precise calibration, additional sensors, or more extensive domain randomization. 
Finally, the time of computation is reasonable (from 5 to 10s), but could be decreased by tuning the architecture of the neural network, lowering the number of optimization steps, or using more powerful hardware. We leave this for future work.

\section{Related work}
\label{sec:related_work}

Over the last decade, progress in multi-fingered robotic grasping has been steady \cite{varley2015generating,lu2019modeling,lu2020planning,lundell2020multi, lundell2021ddgc,wu2020generative, merzic2019leveraging} thanks to differentiable models such as neural networks. Unfortunately, the variety of robotic hands, their actuation modes and sensor inputs, and the lack of standard benchmarks, make it difficult to compare these advances fairly against one another.

From early works, \cite{varley2015generating} identify poses and fingertip positions of stable grasps with a deep neural network from RGB-D images and use a planner (GraspIt!) based on simulated annealing to determine the best hand configurations. They reach a success rate of $75\%$ over a set of 8 objects but suffer from slow execution times (16.6s on average). By contrast, our method is faster and reaches comparable performance over a larger set of objects.
\cite{lundell2020multi} use generative adversarial networks to sample both grasp poses and finger joints efficiently based on RGB-D images. While being a fast approach, they reach only $60\%$ of success rate in real experiments.
The work of \cite{lu2019modeling} is the most similar to ours. They perform grasp planning as probabilistic inference via a classifier trained to predict the success of a grasp. They retrieve maximum a posteriori estimates using gradient ascent and the fact that the classifier is fully differentiable. The prior distribution is fitted with a Gaussian mixture model from the dataset. 
In contrast, our method uses an analytical prior based on power-spherical distributions, does not require an external grasp sampler, and relies on a neural classifier to approximate the likelihood-to-evidence ratio.
Similarly, \cite{lu2020planning} compute maximum likelihood estimates of the hand configuration by making use of gradients provided by a neural network. 
Finally, both of these works treat the rotations with Euler angles and optimize them as real numbers with boundary constraints. 
This representation is not suitable for a neural network according to \cite{zhou2019continuity}.
Instead, our optimization relies on Riemannian conjugate gradients, which preserve the geometrical structure of the rotation group.
Other interesting approaches to multi-fingered grasping include the use of deep reinforcement learning based on vision and tactile sensors~\cite{wu2020generative}, or the use of tactile information only for learning a closed-loop controller~\cite{merzic2019leveraging}.

From a statistical perspective, several Bayesian likelihood-free inference algorithms~\cite{marin2012approximate, beaumont2002approximate, Papamakarios2019SequentialNL, SNPEA, SNPEB, APT, pmlr-v119-hermans20a} have been developed to carry out inference when the likelihood function is implicit and intractable. 
These methods operate by approximating the posterior through rejection sampling or by learning parts of the Bayes' rule, such as the likelihood function, the likelihood-to-evidence ratio, or the posterior itself. 
These algorithms have been used across a wide range of scientific disciplines such as particle physics, neuroscience, biology, or cosmology~\cite{cranmer2020frontier}.
To the best of our knowledge, our work is one of the first to apply one of those for the direct planning successful grasps.
More specifically, we rely here on the amortized inference approach of \cite{pmlr-v119-hermans20a} to carry out inference within seconds for any new observation $i$. In contrast, an approach such as ABC \cite{marin2012approximate, beaumont2002approximate} could take up to hours to determine a single hand configuration $\mathbf{h}$ since data would need to be simulated on-the-fly for each observation $i$ due to the lack of amortization of ABC. 
Neural posterior estimation~\cite{APT} is also amortizable but would have required new methodological developments to be applicable on distributions defined on manifolds, such as those needed here for the rotational part of the pose.


\section{Summary and future work}
\label{sec:summary}

We demonstrate the usefulness and applicability of simulation-based Bayesian inference to multi-fingered grasping.
The approach is generic yet powerful because it can work with any simulator, thereby incorporating from the simplest to the more sophisticated piece of domain knowledge, while leveraging all recent developments from deep learning to solve the Bayesian inference problem. 
Maximum a posteriori hand configurations are found by directly optimizing through the resulting amortized and differentiable expression for the posterior.
The geometry of the configuration space is accounted for by proposing a Riemannian manifold optimization procedure through the neural posterior.
We demonstrate a working proof-of-concept achieving robust multi-fingered grasping, both  in simulation and in real experiments thanks to domain randomization. 
Our success rate is comparable to previous works.


\section*{Acknowledgments}Norman Marlier would like to acknowledge the Belgian Fund for Research training in Industry and Agriculture for its financial support (FRIA grant). Computational resources have been provided by the Consortium des Équipements de Calcul Intensif (CÉCI), funded by the Fonds de la Recherche Scientifique de Belgique (F.R.S.-FNRS) under Grant No. 2.5020.11 and by the Walloon Region.
Gilles Louppe is recipient of the ULiège - NRB Chair on Big data and is thankful for the support of NRB.


\bibliographystyle{unsrt}  
\bibliography{main}  

\newpage
\appendix

\section{Algorithms}
\label{appendix:algorithms}

\IncMargin{1em}
\begin{algorithm}
\DontPrintSemicolon
\SetAlgoNoLine
\KwIn{Priors $p(\mathbf{h})$, $p(\mathcal{O}), p(\mathbf{p}_{\mathcal{O}})$
\newline Sensor generative model $p(i|\mathcal{O}, \mathbf{p}_{\mathcal{O}})$
\newline Grasp generative model $p(S|\mathbf{h}, \mathcal{O}, \mathbf{p}_{\mathcal{O}})$
\newline Criterion $\ell$ (e.g, the binary cross-entropy)}
\KwOut{Trained classifiers $d_{\phi}(S, \mathbf{h})$, $d_{\theta}(i, \mathbf{h})$}
\BlankLine
 \While{not converged}{
  \textbf{Sample} $\mathbf{h} \leftarrow \{\mathbf{h}_{m} \sim p(\mathbf{h})\}_{m=1}^{M}$\;
  \textbf{Sample} $\mathbf{h}' \leftarrow \{\mathbf{h}_{m}' \sim p(\mathbf{h})\}_{m=1}^{M}$\;
  \textbf{Sample} $\mathcal{O}, \mathbf{p}_{\mathcal{O}} \leftarrow \{\mathcal{O}_{m}, \mathbf{p}_{\mathcal{O}, m} \sim p(\mathcal{O})p(\mathbf{p}_{\mathcal{O}})\}_{m=1}^{M}$\;
  \textbf{Simulate} $S \leftarrow \{S_{m} \sim p(S|\mathbf{h}_{m},\mathcal{O}_{m}, \mathbf{p}_{\mathcal{O}, m})\}_{m=1}^{M}$\;
  $\mathcal{L} \leftarrow \ell(d_{\phi}(S, \mathbf{h}), 1) + \ell(d_{\phi}(S, \mathbf{h}'), 0)$\;
  $\phi \leftarrow \texttt{OPTIMIZER}(\phi, \nabla_{\phi}\mathcal{L})$
 }
 \BlankLine
 \While{not converged}{
  \textbf{Sample} $\mathbf{h} \leftarrow \{\mathbf{h}_{m} \sim p(\mathbf{h}|S=1)\}_{m=1}^{M}$\;
  \textbf{Sample} $\mathbf{h}' \leftarrow \{\mathbf{h}_{m}' \sim p(\mathbf{h}|S=1)\}_{m=1}^{M}$\;
  \textbf{Sample} $\mathcal{O}, \mathbf{p}_{\mathcal{O}} \leftarrow \{\mathcal{O}_{m}, \mathbf{p}_{\mathcal{O}, m} \sim p(\mathcal{O},\mathbf{p}_{\mathcal{O}}|S=1,\mathbf{h})\}_{m=1}^{M}$\;
  \textbf{Simulate} $i \leftarrow \{i_{m}\sim p(i|\mathcal{O}_{m}, \mathbf{p}_{\mathcal{O}m})\}_{m=1}^{M}$\;
  $\mathcal{L} \leftarrow \ell(d_{\theta}(i, \mathbf{h}), 1) + \ell(d_{\theta}(i, \mathbf{h}'), 0)$\;
  $\theta \leftarrow \texttt{OPTIMIZER}(\theta, \nabla_{\theta}\mathcal{L})$
 }
 \Return{$d_{\phi}, d_{\theta}$}
 \caption{Training procedure for $d_\phi(S, \mathbf{h})$ and $d_\theta(i, \mathbf{h})$.}
 \label{alg:learning_procedure}
\end{algorithm}
\DecMargin{1em}

\IncMargin{1em}
\begin{algorithm}[H]
\DontPrintSemicolon
\SetAlgoNoLine
\KwIn{Differentiable priors $p(\mathbf{h})$
\newline Differentiable likelihood-to-evidence ratio $\hat{r}$
\newline Depth image $i$}
\KwOut{Approximate MAP estimate $\hat{\mathbf{h}}^{*}$}
\BlankLine
MAP cost function $f(i, \mathbf{h}) = -\log\hat{r}(S=1, i|\mathbf{h})  -\log p(\mathbf{h})$\;
\BlankLine
Sample an initial subset $\mathcal{S} = \{\mathbf{x},\mathbf{q}\}\sim \{ p(\mathbf{x})p(\mathbf{q})\}_{1}^{1000}$\;
\BlankLine
\ForEach{$g_{k} \in \mathcal{G}$}{
$\mathbf{h} = (\mathbf{x}, \mathbf{R}(\mathbf{q}), g_{k})$\;
Initial iterate $\mathbf{h}_{0} = \argmin_{\mathbf{h} \in \mathcal{S}}f(i,\mathbf{h})$\;
$\mathbf{\hat{x}}^{*}_{k}, \mathbf{\hat{R}}^{*}_{k} = \text{Geometric CG method}(f, i,\mathbf{h}_{0})$\;
$\mathbf{\hat{h}}^{*}_{k} = (\mathbf{\hat{x}}^{*}_{k}, \mathbf{\hat{R}}^{*}_{k}, g_{k})$
}
 \Return{$\argmin_{\mathbf{h}_{k}}f(i,\mathbf{\hat{h}}^{*}_{k})$}
 \caption{Manifold optimization procedure to obtain the MAP estimate $\mathbf{\hat{h}}^{*}$ }
 \label{alg:optimization_procedure}
\end{algorithm}
\DecMargin{1em}

\newpage 

\section{Prior Distribution}
\label{appendix:distribution}

\begin{figure}[ht]
     \begin{subfigure}{.5\textwidth}
        \centering
        \includegraphics[width=\textwidth]{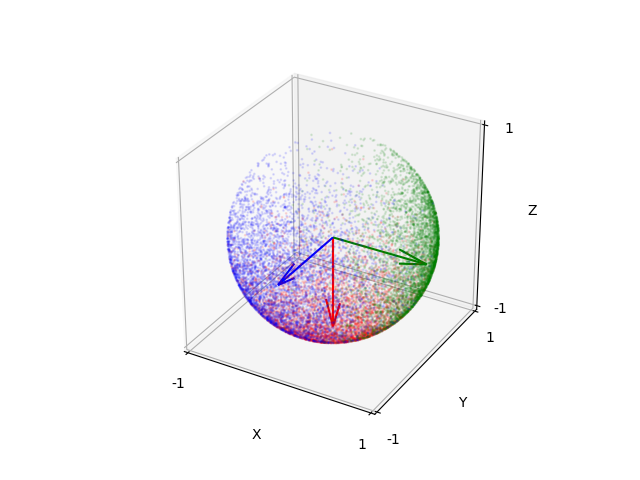}
        \caption{}
        \label{fig:quat_mode1}
     \end{subfigure}
     \hfill
     \begin{subfigure}{.5\textwidth}
        \centering
        \includegraphics[width=\textwidth]{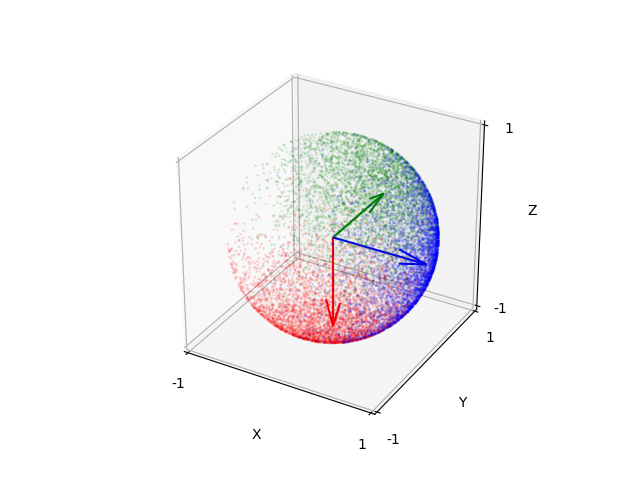}
        \caption{}
        \label{fig:quat_mode2}
     \end{subfigure}
     \vskip\baselineskip
     \begin{subfigure}{.5\textwidth}
        \centering
        \includegraphics[width=\textwidth]{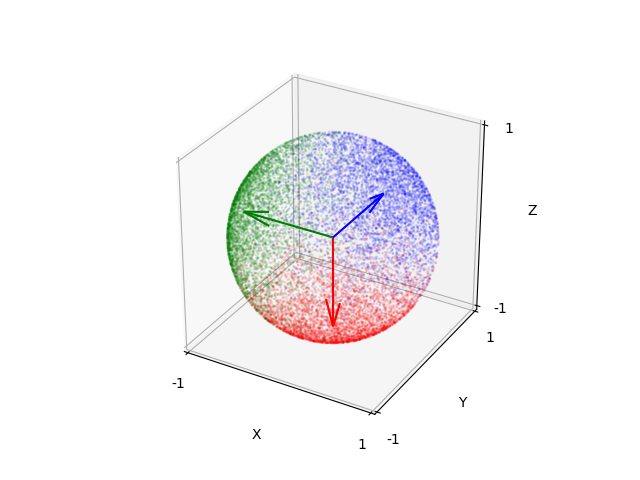}
        \caption{}
        \label{fig:quat_mode3}
     \end{subfigure}
     \hfill
     \begin{subfigure}{.5\textwidth}
        \centering
        \includegraphics[width=\textwidth]{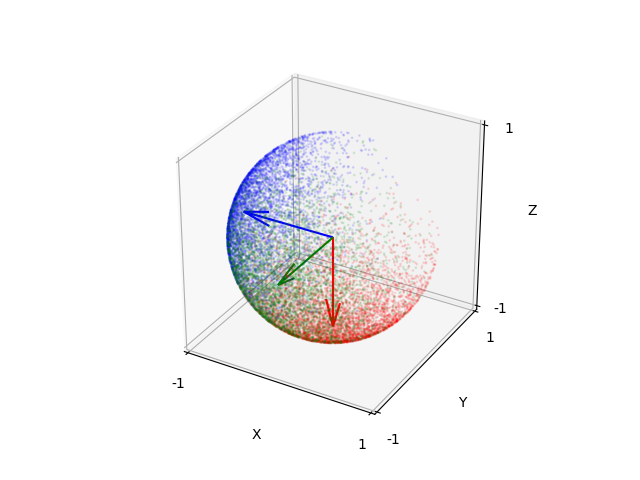}
        \caption{}
        \label{fig:quat_mode4}
     \end{subfigure}
    \caption{Prior distribution $p(\mathbf{q})$. (a)-(d) correspond to the four modes of the mixture.}
    \label{fig:prior_quat_mode}
 \end{figure}

\newpage

\section{YCB objects}
\label{appendix:objects}

\begin{figure}[h]
    \centering
    \includegraphics[scale=0.18]{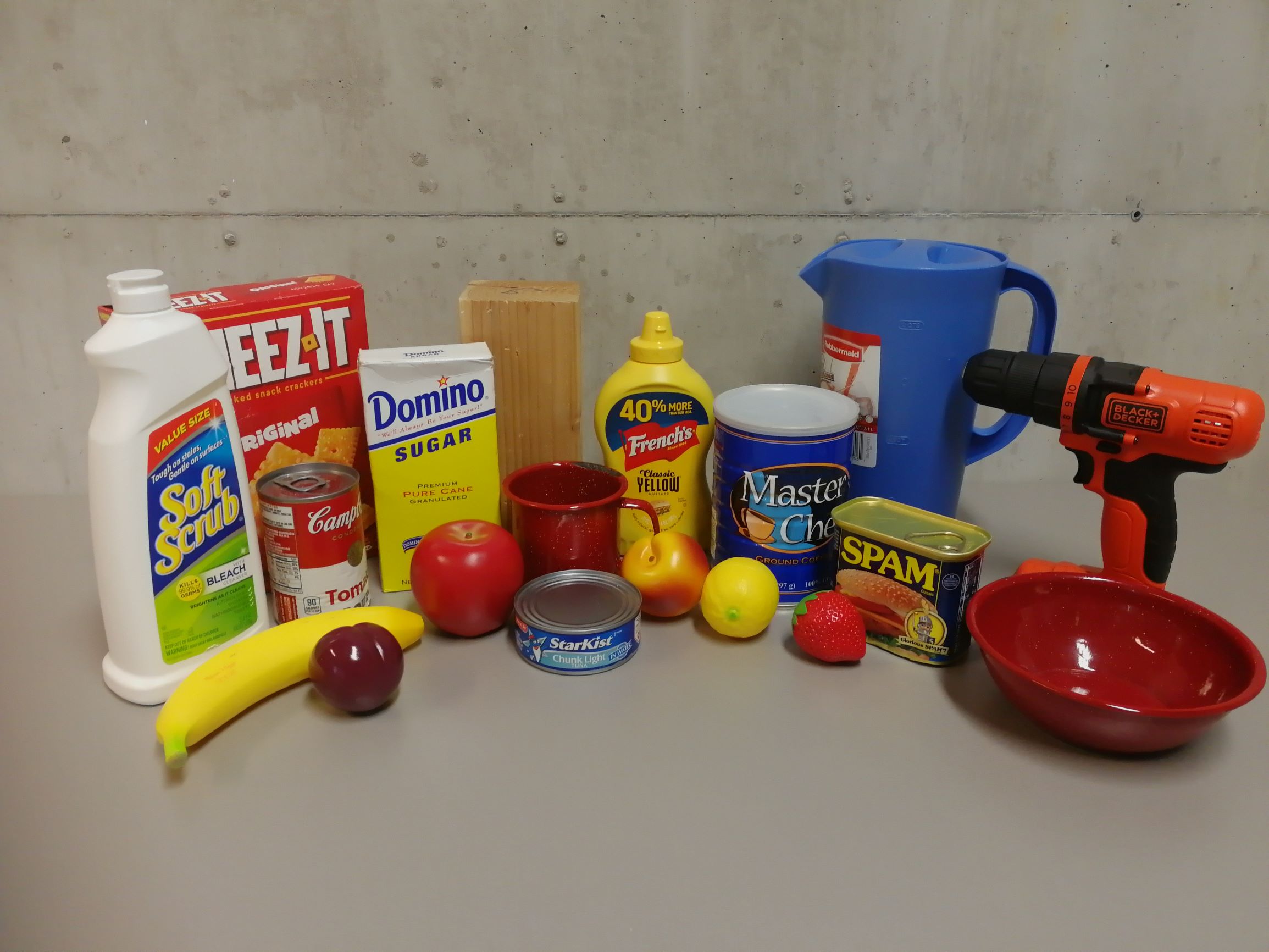}
    \caption{Objects from YCB used for training and testing.}
    \label{fig:training_set}
\end{figure}

\begin{figure}[h]
    \centering
    \includegraphics[scale=0.36]{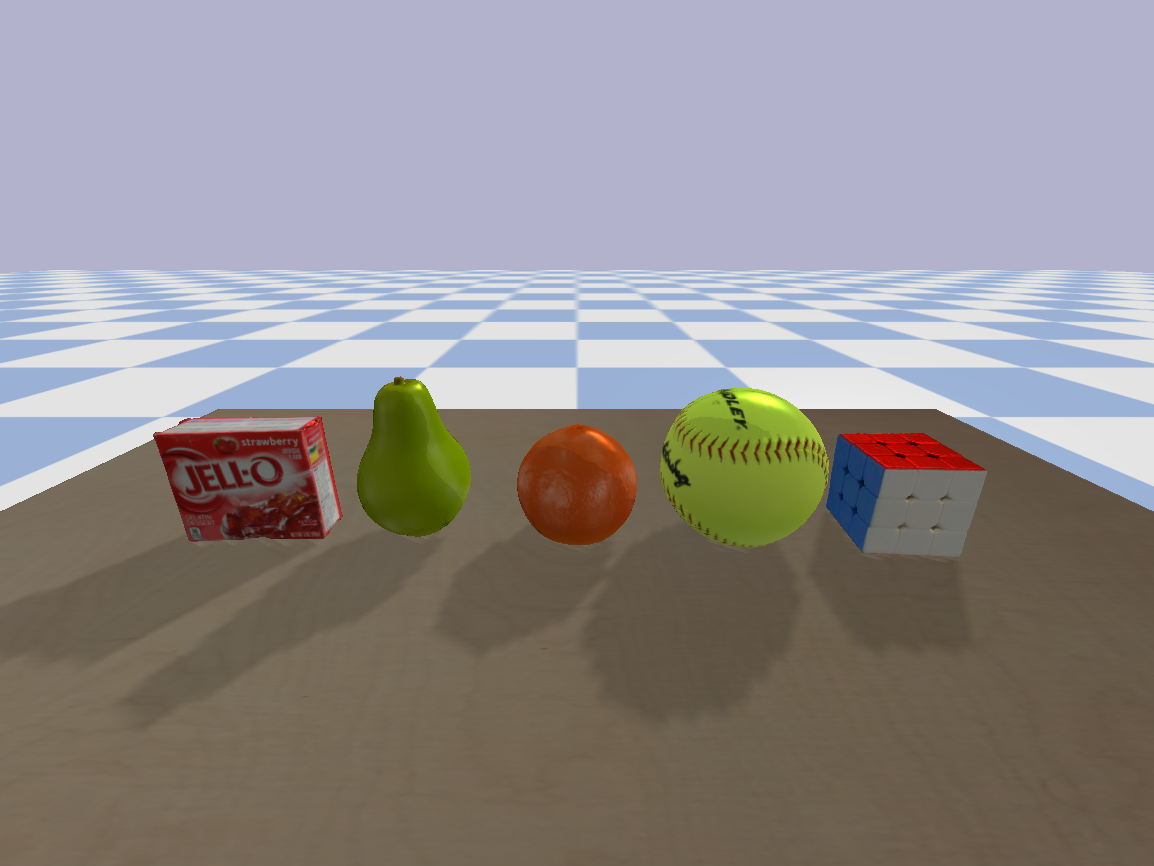}
    \caption{Additional unseen objects from YCB used for testing in simulation.}
    \label{fig:testing_set}
\end{figure}

\newpage

\section{Success rate by object category}
\label{appendix:supplementary-results}

\begin{figure}[h]
    \centering
    \includegraphics[width=\linewidth]{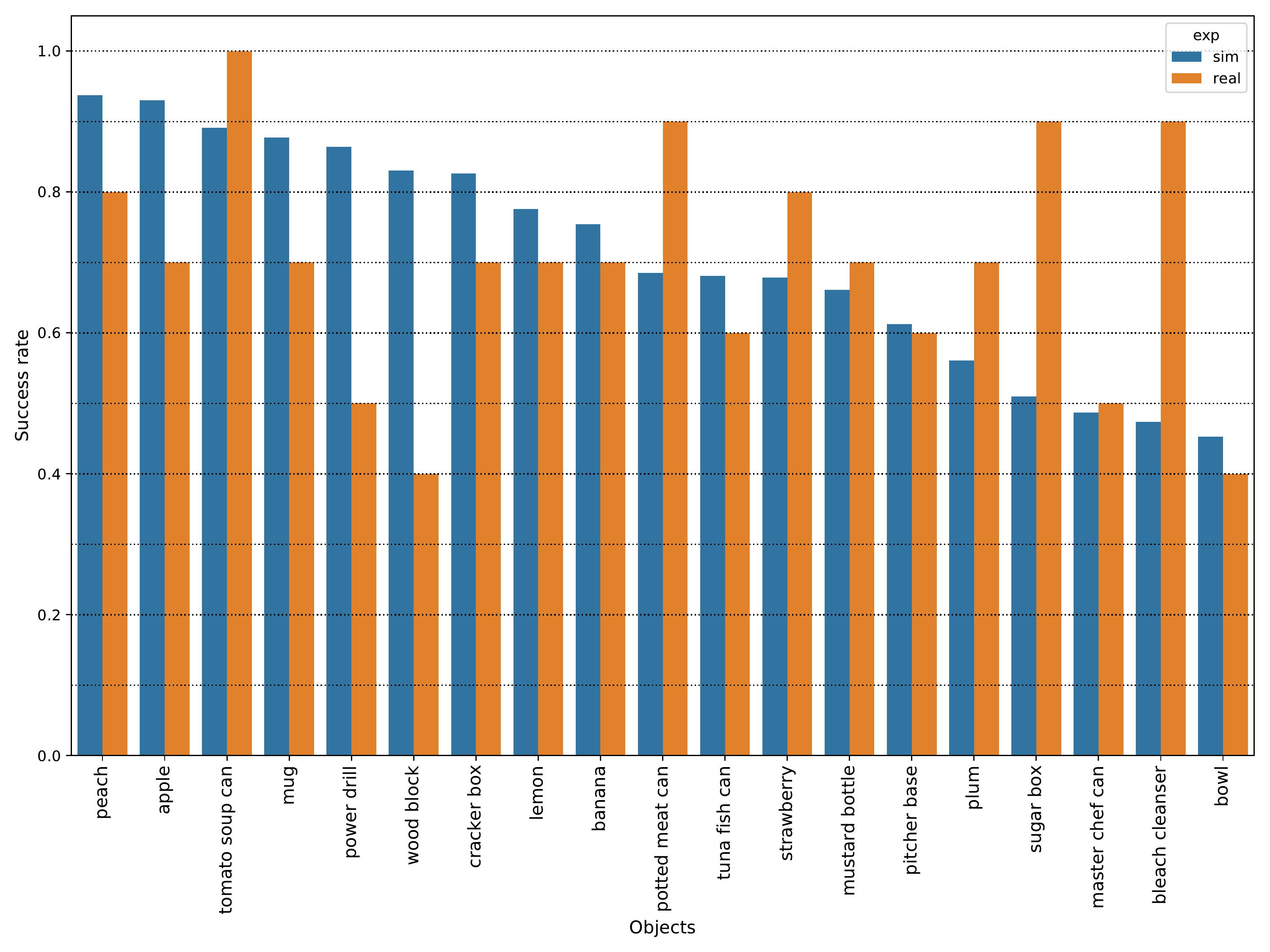}
    \caption{Success rate obtained from simulation and real experiments for each object.}
\end{figure}

\end{document}